\documentclass[pmlr]{jmlr}

\usepackage{longtable}

\usepackage{booktabs}
\usepackage{mathtools}
\usepackage[load-configurations=version-1]{siunitx} 
\usepackage{textcomp}
\makeatletter
\def\set@curr@file#1{\def\@curr@file{#1}} 
\makeatother

\usepackage[normalem]{ulem}

\theorembodyfont{\upshape}
\theoremheaderfont{\scshape}
\theorempostheader{:}
\theoremsep{\newline}

\jmlrvolume{}
\jmlryear{2020}

\title[Towards an Automated SOAP Note]{Towards an Automated SOAP Note: Classifying Utterances from Medical Conversations}

\author{\Name{Benjamin Schloss}
       \Email{bschloss@abridge.com}\\ 
       Abridge AI, Inc.\\
       Pittsburgh, PA, USA 
       \AND
       \Name{Sandeep Konam}
       \Email{san@abridge.com}\\ 
       Abridge AI, Inc.\\
       Pittsburgh, PA, USA} 

\editor{}

\begin{document}

\maketitle

\begin{abstract}

Summaries generated from medical conversations can improve recall and understanding of care plans for patients and reduce documentation burden for doctors. Recent advancements in automatic speech recognition (ASR) and natural language understanding (NLU) offer potential solutions to generate these summaries automatically, but rigorous quantitative baselines for benchmarking research in this domain are lacking. In this paper, we bridge this gap for two tasks: classifying utterances from medical conversations according to (i) the SOAP section and (ii) the speaker role. Both are fundamental building blocks along the path towards an end-to-end, automated SOAP note for medical conversations. We provide details on a dataset that contains human and ASR transcriptions of medical conversations and corresponding machine learning optimized SOAP notes. We then present a systematic analysis in which we adapt an existing deep learning architecture to the two aforementioned tasks. The results suggest that modelling context in a hierarchical manner, which captures both word and utterance level context, yields substantial improvements on both classification tasks. Additionally, we develop and analyze a modular method for adapting our model to ASR output. 
\end{abstract}

\section{Introduction}

Automatically generated summaries of medical conversations between patients and doctors can be useful for both parties. These summaries could help patients improve recall and understanding of their care plan. It is well documented that patients forget 40-80\% of the medical information provided by healthcare practitioners immediately \citep{doi:10.1080/03610739608254020}. Also, patients misconstrue 48\% of what they think they remembered \citep{10.1093/rheumatology/18.1.18}. Increased knowledge and understanding of the details of their care plans may be associated with improved health outcomes \citep{outcomesedu}. In addition, these summaries could help doctors create clinically useful documentation. The clerical burden of entering and maintaining electronic health records (EHRs) is a costly endeavour. By some estimates, doctors may spend as much as two additional hours of administrative work to every one hour spent with patients \citep{sinsky2016allocation}. Increased clerical burden on doctors has been cited as a cause of increasing rates of burnout in the United States in particular \citep{kumar2016burnout}. 

Our approach to this problem is to capture medical conversations as an audio recording, use automatic speech recognition (ASR) to transcribe the conversations, and, then, extract or summarize clinically relevant information with natural language understanding (NLU). Perhaps the most widely used format for clinical notes in the United States is the SOAP note. SOAP stands for Subjective, Objective, Assessment, and Plan, referring to the four major sections of the problem-oriented medical note in which: the patient's reports of symptoms, behaviors, and previous medical conditions are documented (Subjective), the doctor's observations from physical examinations and previously ordered tests are documented (Objective), identification of the potential problem(s), as well as related synthesis of the Subjective and Objective are documented (Assessment), and, finally, the way the problem will be addressed or further investigated is documented (Plan). These sections can be further broken down into subsections. For example, the subjective is often broken into subsections such as the Chief Complaint (the primary reason for the patient's visit), the Review of Systems (a questionnaire of symptoms organized around biological systems), Past Medical History (any past medical problems the patient has had).  


In this paper, we detail the creation of a dataset 
that contains audio recordings of 8,130 medical conversations between doctors and patients, human  and ASR transcriptions of these conversations, and machine learning optimized SOAP notes based on the human generated transcripts. We present a rigorous analysis of a hierarchical encoder decoder architecture \citep{jeblee2019extracting,finley2018automated,liu2019reading}, which we adapted to two basic tasks: classifying utterances from medical conversations (i) according to the SOAP section and (ii) the speaker role (e.g. clinician or patient).  Because ASR transcripts are known to have relatively high numbers of errors compared to human transcripts \citep{kodish2018systematic}, we introduce and analyze a modular technique for testing and training our NLU architecture on ASR generated text. Our results demonstrate that the deep learning architecture examined in this paper outperforms a number of non-deep learning baselines. Furthermore, using bi-LSTMs to contextualize utterance embeddings results in an absolute 6-7\% improvement in $F_{1}$ for SOAP section classification, which reaches near human level performance, an absolute 7-11\% improvement in $F_{1}$ for speaker role classification. Finally, our results also show that our modular approach to ASR adaptation may improve robustness for speaker label but not SOAP section classification.

\subsection*{Generalizable Insights about Machine Learning in the Context of Healthcare}
The ability to identify information in a conversation which is relevant to different parts of the SOAP note and knowing whether a particular utterance was stated by the patient or doctor could be informative to any downstream summarization or information extraction tasks. Thus, we see the two tasks analyzed in this paper as fundamental building blocks in a modular approach to achieving a fully automated SOAP note. Though it is not surprising that dialogues themselves are highly contextual, the relatively large improvement on SOAP section classification from using bi-LSTMs to capture inter-utterance context suggests that a considerable amount of clinically relevant information depends on relatively distant contextual information to be understood. Consequently, simpler extraction methods like string matching, the use of ontologies, sentence parsing, or even using context independent word embedding models will likely result in suboptimal performance because these methods do not capture intersentential context. Additionally, this is the only study to date to provide a rigorous quantitative benchmark for future research on automated SOAP notes for medical conversations, though several research groups have proposed similar architectures and even evaluated similar approaches on limited datasets \citep{jeblee2019extracting,finley2018automated,klann2009intelligent}. Finally, our approach to adapting our model to ASR output provides a generalizable technique for training and evaluating models on ASR data for classification tasks focusing on medical conversations. Compared to other techniques which have primarily focused on joint training and end-to-end solutions \citep{Serdyuk2018TowardsES, lakomkin2019incorporating, morbini2012reranking}, our solution maintains a relatively high degree of modularity which allows it to be used with many open source ASR tools.

\section{Related Work}
\label{related-work}

Previous research has shown it is feasible to generate inferences from structured clinical notes like those found in EHRs \citep{li2017convolutional,guo2019disease,lipton2016learning, mimiciii}. For example, clinical diagnoses have been inferred both from highly structured data, like specific symptoms from a medical ontology \citep{datla2017automated}, as well as from unstructured prose written by a medical professional  \citep{li2017convolutional}. Even more recently, less structured data such as prose has also been automated using summarization techniques \citep{alsentzer2018extractive,liang-etal-2019-novel-system}. 

Improvements have also been made on state-of-the-art ASR models for automatic transcription of clinical dictations and medical conversations between doctors and patients. As of two years ago, word error rates for publicly available ASR systems were somewhere right below 30\% for clinical conversations \citep{kodish2018systematic}. There has also been recent work on improving the performance of publicly available, pretrained ASR systems on medical conversations \citep{mani2020asr,mani2020towards}.
Models specially trained for the medical domain have been able to achieve around 18\% for medical conversations \citep{chiu2018speech} and 16\% for dictations \citep{edwards2017medical}. More recently, \citet{zhou2018analysis} found error rates as low as 7\% for dictated notes. Outside the medical domain, a number of approaches have also been implemented in end-to-end NLU systems which specifically address the problem of ASR errors in NLU tasks. Some learn acoustic embeddings from speech directly \citep{Serdyuk2018TowardsES, palaskar2019learned}, skipping the transcription step, while others use multimodal representations of text and speech \citep{lakomkin2019incorporating}. Still, others have trained ASR and the NLU parts of an end-to-end network jointly instead of separately \citep{morbini2012reranking}. 

These studies suggest that current machine learning techniques are capable of transcribing medical conversations, extracting clinically relevant information, and making clinical inferences of the type that would be necessary to assist both doctors and patients with organizing medical information. However, only a few studies have focused on generating structured summaries of unstructured medical conversations between doctors and patients. One study demonstrated the possibility of extracting symptoms and whether patients confirmed having them or denied having them from human transcribed medical conversations between doctors and patients \citep{rajkomar2019automatically}. Another study provided a systematic analysis of an end-to-end medical scribe, including entity extraction and SOAP classification, but its results are difficult to generalize due to the small dataset and moderate inter-rater reliability \citep{jeblee2019extracting}. Finally, one study has examined classification of medically utterances from a sizeable dataset of medical conversations, but there was no analysis of classification performance on individual SOAP sections \citep{krishna2020extracting}. 

Though previous work has focused on extracting clinically relevant information from clinical narratives \citep{xu2010medex} and biomedical text in general \citep{aronson2001effective,savova2010mayo}, conversational language used in face-to-face dialogues can differ substantially from that found in writing. Only one study to date has incorporated ASR output in the analysis of NLU techniques on medical conversations \citep{selvaraj2019medication}. Other papers have written about end-to-end medical scribes which are in development \citep{finley2018automated,klann2009intelligent}, outlined future research which will include ASR adaptation \citep{jeblee2019extracting}, and improved clinical ASR output through post-processing mechanisms \citep{finley2018dictations}, but the overall lack of meaningful quantitative data in adapting NLU pipelines to ASR transcriptions of medical conversations leaves the possibility of fully automated SOAP annotation an open question.

\begin{table}[ht!]
  \small
  \caption{Data Examples}
  \label{data-examples-table}
  \centering
  \begin{tabular}{ll>{\arraybackslash}p{9cm}}
    \toprule
    \cmidrule(r){1-2}
    SOAP & Speaker & \\
    Section & Label &  Utterance\\
    \midrule
    None & DR &  You are so cool. \\
    None & PT &  Hello. \\
    Subjective & DR & Yeah, the flecainide, you're on 50 twice a day, which obviously a, you know, it's a newer medicine.\\
    Subjective & PT & Because she told me I had a lot of polyps the last time.\\
    Objective & DR & Well, yeah, you did have, according to the last event monitor that I saw you had three episodes of clear AFib.\\
    Objective & PT$^{1}$ & Can we talk about my MRI results? \\
    Assessment & DR & So, I don't know, uh, for the moment I'm assuming that it's related to Parkinson's but in the back of my mind I know it's possible that it's just, uh, its own thing. \\
    Assessment & PT$^{1}$ & So you think my symptoms are being caused by anxiety, not heart problems? \\
    Plan & DR & I want to see you though in four weeks, because I got to do an EKG in four weeks, because there's an area of the heart that we monitor on the EKG when someone is on flecainide to make sure it's not prolonging.\\
    Plan & PT$^{1}$ & Okay, so I should come back in 6 weeks? \\
    \bottomrule
    \multicolumn{3}{p{.85\textwidth}}{Notes. DR = Doctor; PT = Patient; $^{1}$Patient does not typically make objective statements, assessments, or plans, but may solicit relevant information or reiterate what the doctor says.}
  \end{tabular}
\end{table}

\section{Dataset}
\label{dataset}

The data used in this study come from a proprietary dataset consisting of human transcriptions of medical conversations between doctors and patients with speaker labels, the corresponding audio, and human generated SOAP notes based on the transcripts. The entire dataset of human transcripts contains 10,000 hours of transcribed speech (approximately 245,000 unique words and 95,000,000 tokens) and was used to fine-tune a word embedding model and to generate a 5-gram language model. However, we focus on a subset of 1,300 hours of speech for classifying utterances according to the SOAP section and speaker role. This subset of the data consisted of 8,130 doctor patient encounters, including human transcripts, audio, and human generated SOAP notes. For each encounter, we also generate an ASR transcript using the video model from Google's Speech API with automatic diarization and punctuation enabled. Previous work on a related dataset showed that Google's out-of-the-box Speech API achieves around 40 \% WER \citep{mani2020asr}.

A single SOAP note consisted of a set of \emph{observations} which were organized into different SOAP note sections. Each \emph{observation} contained three parts: a short summary of a clinically relevant piece of information (e.g. ``Patient is currently on Plavix"), a set of tags which were specific to each subsection (e.g. ``Plavix", ``current"), and a set of utterances from the doctor patient transcript which were used as evidence for the short summary and tags. The human generated transcripts were broken into utterances by the original transcriptionists. The annotators for this dataset were all highly familiar with medical language and worked in administrative roles in the healthcare system such as transcription of medical conversations and medical billing. In addition, every annotator underwent task-specific training and were subject to both automated and human quality control during actual annotation. One part of this quality control process included having two annotators complete a SOAP note for the same medical conversation and measuring the inter-rater reliability between the two notes. Results from this quality control analysis can be found in Appendix \ref{quality-control-appendix}.

In this study, any utterance which was used as evidence for a note in one of the subsections of the Subjective was classified as \textit{Subjective}, and similarly for the other SOAP sections. Any utterance which was not used as evidence at any point in the SOAP note was classified as \textit{None}. Thus, every utterance in our dataset was associated with a unique speaker label and a SOAP section. Speaker labels were either \textit{Doctor}, \textit{Patient}, \textit{Caregiver} or \textit{Other}. Examples of utterances classified into different SOAP sections can be seen in Table \ref{data-examples-table}, and the distributions of utterances classified into each SOAP section and the distributions of speaker roles can be found in tables \ref{soap-distribution-table} and \ref{speaker-distribution-table}, respectively.

In order to mimic the structure of the human annotations labeled at the utterance level, we restructured the ASR output into utterances, and developed a probabilistic mapping of both the speaker label and the SOAP section from each human transcribed and annotated utterance to utterances reconstructed from the ASR data. This permitted the ASR data to be used during training and testing. The distribution of SOAP sections was highly similar between the original data and the mapped data, but the distribution of speaker labels differed more substantially (see Tables \ref{soap-distribution-table} and \ref{speaker-distribution-table}). A qualitative analysis revealed that systematic diarization related sentence segmentation errors by the ASR system may be the source of this problem. 

\subsection{Mapping Human Annotations to ASR Output}
\label{mapping-human-annottaions-to-ASR-output}

The current approach attempted to map utterance level classifications of SOAP note sections and speaker labels from human transcribed and annotated utterances to utterances that were reconstructed from ASR output. This was achieved by first aligning the human transcript and the ASR output at the character level through a constrained dynamic programming (DP) algorithm. Instead of using DP to align the entire transcripts in a single run, we first searched for sections of the transcript which could be confidently aligned by recursively searching for longest common substrings and partitioning the remaining parts of the transcripts around these substrings. The condition for determining whether to continue partitioning the transcript or to stop depended on the expected n-gram frequency of the longest common substring. If the expected frequency of the n-gram in a given string was less than 0.001, then we considered that n-gram to be unlikely to have occurred purely by chance and continued partitioning. Otherwise, partitioning was stopped and the partition was considered a leaf node. After confidently aligning as many pieces of the two transcripts as possible, we then used DP to align the portions of the transcripts which were in the leaf nodes.

\subsubsection{Word Level Alignment Statistics}
\label{word-level-alignment-statistics}
Once the transcripts were aligned at the character level, we split the aligned ASR output into words and calculated probabilities that a particular word belonged to a certain speaker or SOAP section based on the word(s) to which it was aligned in the human annotated transcript. The probability that a word in the ASR output belonged to a certain class was equal to the percentage of non-space characters in the aligned portion of the human transcript belonging to that class multiplied by confidence in the mapping, which was equal to the number of correctly aligned characters between the ASR word and the aligned portion of the human transcript, divided by the length of the longer of the two segments. This form of label smoothing provides a natural way to capture ASR noise in a continuous fashion and incorporate these effects into the calculation of the loss.

\subsubsection{Sentence Segmentation of ASR Output}
\label{sentence-segmentation-of-ASR-output}
The ASR output used the video model from Google’s speech API with automatic diarization and punctuation enabled. First, we broke the transcript into speaker turns based on the diarization results. Then, we used python’s NLTK library \citep{loper02nltk} to sentence tokenize the speaker turns based on punctuation from the Google API. This resulted in the final set of utterances from the ASR output which were then fed into the model as input for training/testing. 

\begin{table}
  \small
  \caption{SOAP Section Distributions}
  \label{soap-distribution-table}
  \centering
  \begin{tabular}{lcccc}
    \toprule
    \cmidrule(r){1-2}
    SOAP & HT & HT & ASR & ASR\\
    Section & Training & Test & Training & Test \\
    \midrule
    None & 0.63 & 0.64 & 0.61 & 0.63 \\
    Subjective & 0.19 & 0.17 & 0.18 & 0.16 \\ 
    Objective & 0.02 & 0.02 & 0.03 & 0.02 \\
    Assessment & 0.12 & 0.12 & 0.13 & 0.13 \\
    Plan & 0.04 & 0.04 & 0.05 & 0.05 \\
    \bottomrule
    \multicolumn{5}{p{.4\textwidth}}{Notes. HT = Human Transcript.}
  \end{tabular}
\end{table}

\begin{table}
  \small
  \caption{Speaker Label Distributions}
  \label{speaker-distribution-table}
  \centering
  \begin{tabular}{lcccc}
    \toprule
    \cmidrule(r){1-2}
    Speaker & HT & HT & ASR & ASR\\
    Label & Training & Test & Training & Test \\
    \midrule
    Doctor & 0.566 & 0.547 & 0.710 & 0.691 \\
    Patient & 0.383 & 0.391 & 0.264 & 0.277 \\ 
    Caregiver & 0.045 & 0.055 & 0.023 & 0.030  \\
    Other & 0.006 & 0.006 & 0.003 & 0.002 \\
    \bottomrule
    \multicolumn{5}{p{.4\textwidth}}{Notes. HT = Human Transcript.}
  \end{tabular}
\end{table}

\subsubsection{Sentence Level Alignment Statistics}
\label{sentence-level-alignment-stataistics}

\begin{figure*}
    \centering
    \includegraphics[scale=0.275]{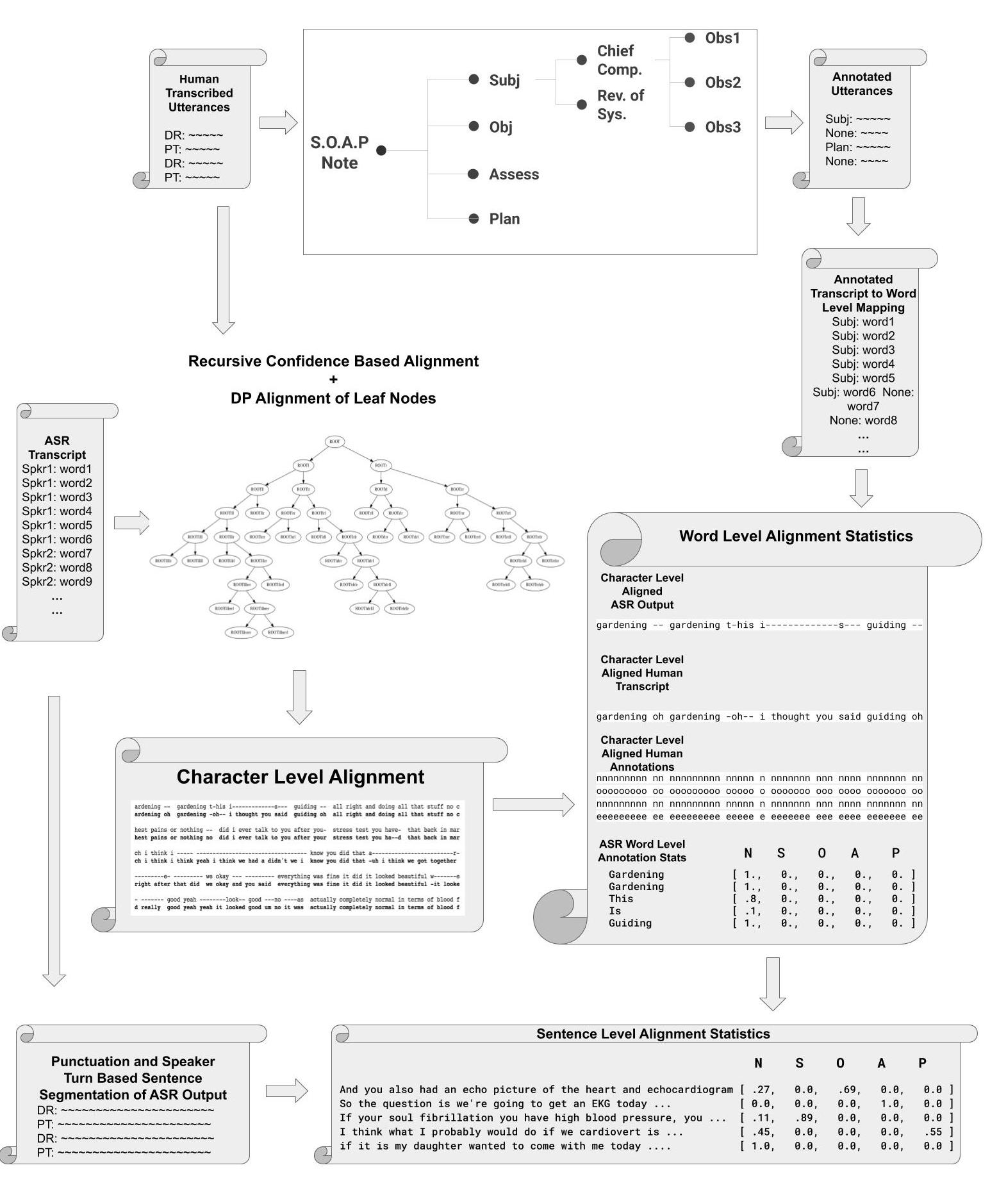}
    \caption{Probabilistic Alignment of Human Annotations to ASR Output}
    \label{alignment-figure}
\end{figure*}

After breaking the ASR output into utterances, each sentence was assigned two probability distributions, one for the five SOAP note classes (including a none class) and another for each of the four speaker classes (Doctor, Patient, Caregiver, or Other) by averaging the probabilities of all the individual words in that utterance (see Section \ref{word-level-alignment-statistics}). Finally, the distributions were normalized to sum to one. Normalization was different for SOAP note sections and speaker labels. For the SOAP note sections, the probability vectors describing each utterance were normalized by adding the difference of one minus the sum of the probabilities of the four SOAP sections to the None section. This normalization method was motivated by the fact that, if the ASR mapping is poor, then it is likely that the content of the ASR transcript in a particular utterance does not closely match the human annotated transcript. Since the SOAP note section classification is primarily based on the content of the utterance, changing the content should also change the classification. For example, if the original utterance was ``We're going to start you on baby aspirin" and was classified as \textit{Plan}, and the corresponding ASR output was ``We're going to start you baby as per him," we may want to hedge our confidence in classifying the latter utterance as part of the \textit{Plan} because, depending on the degradation of the ASR output, the content might not actually have any useful information pertaining to the Plan.  On the other hand, for the speaker, we did not allow for an \textit{Unknown} option during classification, and, although the content may have differed between the ASR and human transcript, this would not fundamentally change who is speaking at a certain time. For this reason, the probability vectors for the speaker class labels were normalized by simply dividing by the L2 norm of the vector. 

\subsubsection{Data Preprocessing}
\label{data-preprocessing}
Each non-empty utterance which contained at least one word (was not just laughter or some other non-linguistic sound) was extracted from the transcript. Sentences were split using Python's NLTK library and non-word annotations (primarily non-linguistic sounds or redacted content for deidentification of protected health information) were standardized to a single unique word in capital letters (e.g. [physician name]\textrightarrow PHYSICIAN\_NAME). Dashes at the end of utterances were removed. Finally, for utterances which were greater than 32 tokens after splitting, we removed all stop words, and, if there were still more than 32 tokens after removing stop words, we took only the first 32 tokens of the utterance. Utterances shorter than 32 tokens were padded with empty strings.

\section{Methods}
\label{traditional-machine-learning-and-deep-learning-models}

We assess NLU for medical conversations using both bag-of-words (BoW) and deep learning models in this paper. For the BoW baselines, we use majority-class (MC), Multinomial Naive Bayes (MNB), logistic regression (LR), and random forest (RF) classifiers with a BoW encoding of each utterance. Building off of \citet{jeblee2019extracting}, \citet{finley2018automated}, and \citet{liu2019reading}, we explore four gradually more complex deep learning models. The most basic model, which we refer to as the deep learning baseline (DLB), learns a weighted average of the three ELMo layers and averages all of the word embeddings to generate a sentence embedding \citep{peters2018deep}. ELMo is a contextualized  word embedding model, meaning it integrates sentence context into representations of word meaning. Then, each model after that adds one specific layer. First, we add word attention to generate a word weighted utterance embedding. Second, we add a bi-LSTM to generate contextualized utterance embeddings, and, lastly, we add an LSTM decoder to generate contextualized outputs. All models ended with a dense layer followed by a softmax activation function used for classification. 

The full encoder produces a contextualized utterance, $\mathbf{c_i}; i \in \mathbb{N}$ by first generating ELMo representations from all tokens $\mathbf{t_{ij}}$ using a pretrained ELMo model with 512 hidden units,  $\mathbf{e_{ikj}}$ with 3 layers ($k \in \{1,2,3\}$) each containing 1024 units per layer after concatenating both directions (resulting in a 3x1024 matrix). ELMo had been pretrained by fine-tuning the out-of-the-box, large ELMo model for three epochs on all of the human transcribed utterances from our proprietary dataset. An attention layer, $\mathbf{a_{L_i}}$, was used to combine the three ELMo layers, resulting in a single 1x1024 dimensional embedding for each token, $\mathbf{l_{ij}}$, in an utterance. Utterances were zero padded to 32 tokens each or truncated to 32 tokens, $j \in \{1...32\}$ (see Section \ref{data-preprocessing} for more details). Then, a word attention layer, $\mathbf{a_{w_i}}$, was used to create a weighted average of the word embeddings in an utterance, resulting in a final 1x1024 dimensional utterance embedding, $\mathbf{u_i}$. Attention weights for both ELMo layer level and word level attention were scaled using softmax before combining representations. These utterance embeddings were then passed to a two layer stacked bi-LSTM in which layers one and two had 512 and 256 dimensions (1024 and 512 after concatenation), respectively. The output of the final layer resulted in a 512 dimensional, contextualized encoding of each utterance, $\mathbf{c_i}$. A dropout layer was added for the input of the bi-LSTM layers and for the recurrent layers. 


\begin{align*}
    \phantom{\mathbf{a_{L_{ij}}}}
    &\begin{aligned}
      \mathllap{E_{ij}} &= ELMo(\mathbf{t_{ij}});\ j \in \{1...32\},\ i \in \mathbb{N}
    \end{aligned}\\
    &\begin{aligned}
      \mathllap{\mathbf{a_{L_{ij}}}} &=  softmax([dot(\mathbf{w_{L}},\  \mathbf{e_{i1j}})\ ...\ dot(\mathbf{w_{L}},\  \mathbf{e_{i3j}})]);\ j \in \{1...32\},\ i \in \mathbb{N}
    \end{aligned}\\
    &\begin{aligned}
      \mathllap{\mathbf{l_{ij}}} &=  \sum^3_{k}a_{L_{ikj}} \mathbf{e_{ikj}};\ j \in \{1...32\},\ i \in \mathbb{N}
    \end{aligned}\\
    &\begin{aligned}
      \mathllap{\mathbf{a_{w_{i}}}} &=  softmax([dot(\mathbf{w_w},\  \mathbf{l_{i1}})\ ...\ dot(\mathbf{w_w},\ \mathbf{l_{i32}})]);\ i \in \mathbb{N}
    \end{aligned}\\
    &\begin{aligned}
      \mathllap{\mathbf{u_i}} &=  \sum^{32}_{j}a_{w_{ij}} \mathbf{l_{ij}};\ i \in \mathbb{N}
    \end{aligned}\\
    &\begin{aligned}
      \mathllap{\mathbf{c_i}} &=  StackedBiLSTM(\mathbf{u_i})\ i \in \mathbb{N}
    \end{aligned}
\end{align*}

A multi-decoder consisted of two separate unidirectional LSTMs which decoded the speaker labels and the SOAP note sections separately. The reason for designing a multi-decoder and framing the problem as a multitask problem, which decodes the speaker and the SOAP section, is that some of the SOAP note sections are defined differently with respect to the doctor and the patient (see Table \ref{data-examples-table} and Section \ref{dataset} for more details). Namely, the Subjective is supposed to capture the patient's reports and experience--but may be directed by the doctor's questions--while the Objective, Assessment, and Plan are typically driven by the doctor's own measurements, observations, and reasoning. Therefore, we weighted the loss function by its ability to solve both problems simultaneously. A dropout layer was added for the input of the decoder LSTMs and for the recurrent layers. At each step, the contextualized encoding, $\mathbf{c_i}$, for the corresponding utterance was the input. For the first utterance in each transcript, the initial states of the first decoding steps were set to all zeros. The speaker decoder had 64 hidden units and 4 projection units (one for each speaker class), and the SOAP section decoder had 64 hidden units and 5 projection units (one for each section class). The projection layer representations were then passed to a softmax layer to generate the final sequence of speaker and section probabilities, $\{\mathbf{s_{pkr_i}}\}_{i \in N}$ and $\{\mathbf{s_{ect_i}}\}_{i \in N}$, given a contextualized encoding of an input token sequence $\mathbf{c_i}$ and all preceding contextualized encodings, $\{\mathbf{c_1},...\mathbf{c_{i-1}}\}$. 

\begin{align*}
    \phantom{P(\mathbf{s_{ect_i}}|\mathbf{c_i},\mathbf{c_1},...,\mathbf{c_{i-1}})}
    &\begin{aligned}
      \mathllap{P(\mathbf{s_{pkr_i}}|\mathbf{c_i},\mathbf{c_1},...\mathbf{c_{i-1}})} &= softmax(\mathbf{w_{spkr}}LSTM_{spkr}(\mathbf{c_i}) + \mathbf{b_{spkr}})
    \end{aligned}\\
    &\begin{aligned}
      \mathllap{P(\mathbf{s_{ect_i}}|\mathbf{c_i},\mathbf{c_1},...,\mathbf{c_{i-1}})} &= softmax(\mathbf{w_{sect}}LSTM_{sect}(\mathbf{c_i}) + \mathbf{b_{sect}})
    \end{aligned}
\end{align*}

\subsection{Model Training}
\label{model-training}

The deep learning models were trained  using a weighted cross entropy loss function with an ADAM optimizer; the learning rate was set to 0.001. Each model was trained in batches of 4 transcripts at a time, the order of which was randomized during each epoch. The truncated back propagation length was set to 64 for all LSTMs. During training, the dropout rate decreased over a period of 5 epochs, taking the following values for each epoch: 0.45, 0.30, 0.25, 0.22, and 0.21.  During test, the dropout rate was set to 0. No other regularization was used. Gradients were clipped using a scale factor of 5 of the global norm. The loss for each batch was the sum of the loss for the speaker classification and the SOAP section classification. The loss was weighted by inverse class-frequency per batch. Hyperparameters were based on results from Bayesian hyperparameter optimization. ELMo weights were frozen during training. 

The MNB classifier was trained using a uniform prior for the classes, and the LR and RF classifiers were trained by weighting the loss function based on class frequency (LR and RF). Thus, the training for the traditional machine learning baselines was comparable to the class frequency weighting of the loss function during training of the deep learning models. Hyperparameters for the BoW models were not based on Bayesian hyperparameter optimization.

All models were trained once using only utterances from the human transcripts (labeled as ``ASR" in subsequent tables) and once using both utterances from the ASR output and the human transcripts (labeled as ``w/ ASR" in subsequent tables). In turn, both training methods were evaluated at test on both ASR and human transcribed (HT) utterances separately so that the effect of including ASR in training could be evaluated on both ASR and human transcribed data and compared to training without ASR data. In addition, we explored the effect of calibrating models using Platt scaling and a 10\% validation set. 

The losses for the ASR utterances and the utterances for the human transcripts were slightly different. The expected labels for the human transcripts were one-hot encoded; however, the expected labels for the ASR utterances reflected the noise in the mapping, and were probability distributions of each class inferred by the mapping. This is effectively a form of label smoothing for the noise introduced by the ASR transcript. See Section \ref{sentence-level-alignment-stataistics} for more details.

\section{Results}
\label{results}

\subsection{Model Evaluation}
\label{model-evaluation}

For this analysis, we use accuracy, $F_{1}$ (macro for multiclass, standard $F_{1}$ otherwise), the area under the curve of the receiver operating characteristic (AUROC) curve, and area under the precision-recall curve (AUPRC) to evaluate the classification performance of our models. We also assess how including or excluding ASR data in training using the mapping described in \ref{mapping-human-annottaions-to-ASR-output} affects model performance by training all of our models with and without ASR data and evaluating on human transcribed and ASR data separately. Finally, we assess how calibrating our models using a 10\% validation set and Platt scaling affects model performance across different metrics.

\subsection{SOAP and Speaker Classification Results}
\label{soap-and-speaker-classification-results}

\begin{table}
  \footnotesize
  \caption{SOAP Classification as a Function of Model Type and Training Data}
  \label{clf-SOAP-section-performance-table}
  \centering
  \begin{tabular}{llcc|cc|cc|cc}
    \toprule
    \cmidrule(r){1-2}
    & & Acc & Acc & $F_{1}$ & $F_{1}$ & AUROC & AUROC & AUPRC & AUPRC \\
    & & (train & (train & (train & (train & (train & (train & (train & (train \\
    Test & & w/o & w/ & w/o & w/ & w/o & w/ & w/o & w/ \\
    Set & Model & ASR) & ASR) & ASR) & ASR) & ASR) & ASR) & ASR) & ASR) \\
    \midrule
    HT & MC & 0.64 & 0.64 & 0.16 & 0.16 & 0.50 & 0.50 & 0.20 & 0.20  \\
    HT & BoW+NB & 0.66 & 0.66 & 0.36* & 0.36* & 0.71 & 0.71* & 0.37 & 0.37 \\
    HT & BoW+LR & 0.66 & 0.66 & 0.27* & 0.35* & 0.70 & 0.70 & 0.35 & 0.35 \\
    HT & BoW+RF & 0.64 & 0.64 & 0.29* & 0.29* & 0.65* & 0.65* & 0.30 & 0.30 \\
    HT & DLB & 0.67 & 0.66 & 0.40* & 0.40* & 0.76* & 0.75* & 0.39 & 0.38 \\
    HT & DLB+WA & 0.67 & 0.67 & 0.41* & 0.40* & 0.76* & 0.76* & 0.41 & 0.40 \\
    HT & DLB+WA+BiL &  \textbf{0.68} & \textbf{0.68} & 0.47 & \textbf{0.48} & \textbf{0.86*} & \textbf{0.86*} & \textbf{0.50} & \textbf{0.50} \\
    HT & DLB+WA+BiL+LD & \textbf{0.68} & \textbf{0.68} & 0.47 & \textbf{0.48} & \textbf{0.86*} & \textbf{0.86*} & \textbf{0.50} & \textbf{0.50} \\ 
    \midrule
    ASR & MC & 0.63 & 0.63 & 0.15 & 0.15 & 0.50 & 0.50 & 0.20 & 0.20 \\
    ASR & BoW+NB & 0.64 & 0.64 & 0.32* & 0.33* & 0.70 & 0.70 & 0.35 & 0.35 \\
    ASR & BoW+LR & 0.63 & 0.64 & 0.27* & 0.31* & 0.69* & 0.70* & 0.33  & 0.34 \\
    ASR & BoW+RF & 0.63 & 0.63 & 0.28* & 0.28* & 0.65* & 0.65* & 0.29 & 0.29 \\
    ASR & DLB & 0.64 & 0.64 & 0.38* & 0.38* & 0.74* & 0.73* & 0.36 & 0.35 \\
    ASR & DLB+WA & 0.64 & 0.64 & 0.38* & 0.39* & 0.74* & 0.74* & 0.37 & 0.36 \\
    ASR & DLB+WA+BiL & 0.64 & 0.64 & 0.42* & \textbf{0.45*} & 0.82* & \textbf{0.83*} & 0.43 & \textbf{0.44} \\
    ASR & DLB+WA+BiL+LD & \textbf{0.65}  & 0.64 & 0.42* & 0.44* & \textbf{0.83*} &
    \textbf{0.83*} & \textbf{0.44} & \textbf{0.44}\\
    \bottomrule
    \multicolumn{10}{p{.99\textwidth}}{Notes. HT = Human Transcript; BoW = Bag of Words; DLB = Deep Learning Baseline (ELMo + Layer Attention + Average Word Embedding); WA = Word Attention; LD = LSTM Decoder; All models ended with a dense layer followed by a softmax activation function. Cases in which the uncalibrated model outperformed the calibrated model are marked with an asterisk (*).} 
  \end{tabular}
\end{table}

\begin{table}
  \footnotesize
  \caption{Speaker Classification as a Function of Model Type and Training Data}
  \label{clf-speaker-label-performance-table}
  \centering
  \begin{tabular}{llcc|cc|cc|cc}
    \toprule
    \cmidrule(r){1-2}
    & & Acc & Acc & $F_{1}$ & $F_{1}$ & AUROC & AUROC & AUPRC & AUPRC \\
    & & (train & (train & (train & (train & (train & (train & (train & (train \\
    Test & & w/o & w/ & w/o & w/ & w/o & w/ & w/o & w/ \\
    Set & Model & ASR) & ASR) & ASR) & ASR) & ASR) & ASR) & ASR) & ASR) \\
    \midrule
    HT & MC & 0.55 & 0.55 & 0.18 & 0.18 & 0.50 & 0.50 & 0.25 & 0.25 \\
    HT & BoW+NB & 0.65 & 0.65 & 0.38* & 0.38* & 0.72* & 0.71 & 0.40 & 0.40 \\
    HT & BoW+LR & 0.61 & 0.60 & 0.34* & 0.29* & 0.68* & 0.70* & 0.37 & 0.36 \\
    HT & BoW+RF & 0.60 & 0.58 & 0.32* & 0.29* & 0.69* & 0.68* & 0.37 & 0.36 \\
    HT & DLB & 0.70 & 0.70 & 0.42* & 0.42* & 0.75* & 0.75* & 0.37 & 0.36\\
    HT & DLB+WA & 0.70 & 0.69 & 0.42* & 0.42* & 0.76* & 0.75* & 0.37 & 0.37 \\
    HT & DLB+WA+BiL & \textbf{0.82} & \textbf{0.82} & \textbf{0.59*} & 0.58 & \textbf{0.92*} & 0.91* & \textbf{0.51*} & \textbf{0.51}\\
    HT & DLB+WA+BiL+LD & 0.81 & 0.81 & \textbf{0.59*} & 0.56* & 0.91* & 0.91* & 0.5 & 0.49 \\
    \midrule
    ASR & MC & 0.69 & 0.69 & 0.20 & 0.20 & 0.50 & 0.50 & 0.25 & 0.25 \\
    ASR & BoW+NB & 0.71 & 0.71 & 0.33 & 0.33* & 0.70 & 0.70 & 0.36 & 0.37 \\
    ASR & BoW+LR & 0.67 & 0.71 & 0.31 & 0.30 & 0.66 & 0.66 & 0.35 & 0.35 \\
    ASR & BoW+RF & 0.69 & 0.71 & 0.29* & 0.29* & 0.67* & 0.67* & 0.34* & 0.34 \\
    ASR & DLB & 0.73 & 0.75 & 0.37 & 0.39* & 0.72* & 0.73* & 0.33 & 0.33 \\
    ASR & DLB+WA & 0.73 & 0.76 & 0.37 & 0.39* & 0.73* & 0.73* & 0.34 & 0.33 \\
    ASR & DLB+WA+BiL & 0.70 & 0.79 & 0.42 & \textbf{0.50*} & 0.82* & 0.85* & 0.35 & \textbf{0.42} \\
    ASR & DLB+WA+BiL+LD & 0.71 & \textbf{0.80} & 0.42 & 0.49* & 0.82* & \textbf{0.86*} & 0.36 & 0.41 \\
    \bottomrule
    \multicolumn{10}{p{.99\textwidth}}{Notes. HT = Human Transcript; BoW = Bag of Words; DLB = Deep Learning Baseline (ELMo + Layer Attention + Average Word Embedding); WA = Word Attention; LD = LSTM Decoder; All models ended with a dense layer followed by a softmax activation function. Cases in which the uncalibrated model outperformed the calibrated model are marked with an asterisk (*).} 
  \end{tabular}
\end{table}

\begin{table}
  \footnotesize
  \caption{SOAP Classification for Deep Learning Models by SOAP Section}
  \label{soap-section-table}
  \centering
  \begin{tabular}{>{\arraybackslash}p{1.5cm}>{\arraybackslash}p{0.75cm}>{\arraybackslash}p{1.5cm}>{\centering\arraybackslash}p{1.75cm}>{\centering\arraybackslash}p{1.75cm}>{\centering\arraybackslash}p{2.5cm}>{\centering\arraybackslash}p{2.5cm}}
    \toprule
    \cmidrule(r){1-2}
    SOAP & Test & & & & & \\
    Section & Set & Training & DLB & + WA & + Bi-LSTM & + LD \\
    \midrule
    None & HT & w/o ASR & 0.79 [0.71] & \textbf{0.80} [0.71] & 0.79 [0.67] & 0.79 [0.66] \\
    None & HT & w/ ASR & \textbf{0.79} [0.73] & \textbf{0.79} [0.72] & \textbf{0.79} [0.66] & \textbf{0.79} [0.66] \\
    None & ASR & w/o ASR & \textbf{0.78} [0.67] & \textbf{0.78} [0.67] & 0.77 [0.61] & 0.77 [0.62] \\
    None & ASR & w/ ASR & \textbf{0.78} [0.74] & \textbf{0.78} [0.72] & \textbf{0.78} [0.70] & \textbf{0.78} [0.69] \\
    Subjective & HT & w/o ASR & 0.32 [0.42] & 0.34 [0.42] & 0.49 [\textbf{0.52}] & 0.49 [\textbf{0.52}] \\
    Subjective & HT & w/ ASR & 0.30 [0.41] & 0.33 [0.42] & 0.50 [\textbf{0.52}] & 0.50 [\textbf{0.52}] \\
    Subjective & ASR & w/o ASR & 0.25 [0.36] & 0.25 [0.36] & 0.38 [0.44] & 0.36 [\textbf{0.45}] \\
    Subjective & ASR & w/ ASR & 0.19 [0.34] & 0.20 [0.35] & 0.25 [0.44] & 0.19 [\textbf{0.45}] \\
    Objective & HT & w/o ASR & 0.33 [0.28] & 0.34 [0.30] & \textbf{0.49} [0.38] & \textbf{0.49} [0.40] \\
    Objective & HT & w/ ASR & 0.31 [0.28] & 0.33 [0.29] & 0.48 [0.39] & \textbf{0.50} [0.38] \\
    Objective & ASR & w/o ASR & 0.27 [0.25] & 0.28 [0.26] & \textbf{0.40} [0.36] & 0.39 [0.37] \\
    Objective & ASR & w/ ASR & 0.22 [0.26] & 0.25 [0.27] & 0.14 [\textbf{0.38}] & 0.04 [0.36] \\
    Assessment & HT & w/o ASR & 0.13 [0.28] & 0.15 [0.30] & 0.24 [\textbf{0.37}] & 0.25 [\textbf{0.37}] \\
    Assessment & HT & w/ ASR & 0.13 [0.26] & 0.14 [0.28] & 0.29 [\textbf{0.37}] & 0.28 [\textbf{0.37}] \\
    Assessment & ASR & w/o ASR & 0.17 [0.31] & 0.17 [0.32] & 0.22 [\textbf{0.36}] & 0.28 [\textbf{0.36}] \\
    Assessment & ASR & w/ ASR & 0.08 [0.26] & 0.09 [0.29] & 0.06 [\textbf{0.36}] & 0.04 [\textbf{0.36}] \\
    Plan & HT & w/o ASR & 0.24 [0.31] & 0.25 [0.31] & 0.34 [\textbf{0.36}] & 0.32 [0.35] \\
    Plan & HT & w/ ASR & 0.22 [0.30] & 0.24 [0.30] & 0.34 [\textbf{0.37}] & 0.35 [0.36] \\
    Plan & ASR & w/o ASR & 0.21 [0.29] & 0.21 [0.29] & 0.28 [0.32] & 0.25 [\textbf{0.33}] \\
    Plan & ASR & w/ ASR & 0.16 [0.29] & 0.18 [0.29] & 0.15 [0.34] & 0.13 [\textbf{0.35}] \\
    \bottomrule
    \multicolumn{7}{p{.95\textwidth}}{Notes. HT = Human Transcript; DLB = Deep Learning Baseline (ELMo + Layer Attention + Average Word Embedding); WA = Word Attention; LD = LSTM Decoder; All models ended with a dense layer followed by Softmax. All scores are single class $F_{1}$ scores. Uncalibrated model results are presented in brackets.}
  \end{tabular}
\end{table}

\begin{table}
  \footnotesize
  \caption{Speaker Classification for Deep Learning Models by Speaker Label}
  \label{speaker-label-table}
  \centering
  \begin{tabular}{>{\arraybackslash}p{1.5cm}>{\arraybackslash}p{0.75cm}>{\arraybackslash}p{1.5cm}>{\centering\arraybackslash}p{1.75cm}>{\centering\arraybackslash}p{1.75cm}>{\centering\arraybackslash}p{2.5cm}>{\centering\arraybackslash}p{2.5cm}}
    \toprule
    \cmidrule(r){1-2}
    Speaker & Test & & & & & \\
    Label & Set & Training & DLB & + WA & + Bi-LSTM & + LD \\
    \midrule
    Doctor & HT & w/o ASR & 0.76 [0.72] & 0.75 [0.70] & \textbf{0.86} [0.85] & \textbf{0.86} [0.84] \\
    Doctor & HT & w/ ASR & 0.76 [0.72] & 0.76 [0.72] & \textbf{0.86} [0.83] & 0.85 [0.82] \\
    Doctor & ASR & w/o ASR & \textbf{0.82} [0.73] & 0.81 [0.72] & 0.79 [0.74] & 0.79 [0.74] \\
    Doctor & ASR & w/ ASR & 0.84 [0.79] & 0.84 [0.78] & \textbf{0.86} [0.83] & \textbf{0.86} [0.83] \\
    Patient & HT & w/o ASR & 0.67 [0.67] & 0.67 [0.67] & \textbf{0.81} [\textbf{0.81}] & 0.80 [0.80] \\
    Patient & HT & w/ ASR & 0.65 [0.68] & 0.62 [0.67] & \textbf{0.81} [0.79] & 0.80 [0.79] \\
    Patient & ASR & w/o ASR & 0.58 [0.56] & 0.58 [0.56] & 0.58 [\textbf{0.59}] & \textbf{0.59} [\textbf{0.59}] \\
    Patient & ASR & w/ ASR & 0.55 [0.58] & 0.55 [0.57] & 0.65 [0.66] & 0.65 [\textbf{0.67}] \\
    Caregiver & HT & w/o ASR & 0.08 [0.23] & 0.09 [0.23] & 0.40 [\textbf{0.46}] & 0.42 [0.44] \\
    Caregiver & HT & w/ ASR & 0.06 [0.21] & 0.07 [0.24] & 0.25 [0.42] & 0.30 [\textbf{0.46}] \\
    Caregiver & ASR & w/o ASR & 0.08 [0.14] & 0.07 [0.14] & 0.25 [0.23] & \textbf{0.30} [0.24] \\
    Caregiver & ASR & w/ ASR & 0.05 [0.17] & 0.04 [0.17] & 0.28 [0.33] & 0.18 [\textbf{0.36}] \\
    Other & HT & w/o ASR & 0.00 [0.07] & 0.00 [0.08] & 0.23 [0.26] & 0.05 [\textbf{0.29}]  \\
    Other & HT & w/ ASR & 0.00 [0.07] & 0.00 [0.07] & 0.20 [\textbf{0.24}] & 0.00 [0.18] \\
    Other & ASR & w/o ASR & 0.00 [0.01] & 0.00 [0.02] & 0.05 [0.06] & 0.00 [\textbf{0.08}] \\
    Other & ASR & w/ ASR & 0.00 [0.02] & 0.00 [0.03] & 0.01 [\textbf{0.16}] & 0.00 [0.11] \\
    \bottomrule
    \multicolumn{7}{p{.95\textwidth}}{Notes. HT = Human Transcript; DLB = Deep Learning Baseline (ELMo + Layer Attention + Average Word Embedding); WA = Word Attention; LD = LSTM Decoder; All models ended with a dense layer followed by Softmax. All scores are single class $F_{1}$ scores. Uncalibrated model results are presented in brackets.}
  \end{tabular}
\end{table}

Tables \ref{clf-SOAP-section-performance-table} and \ref{clf-speaker-label-performance-table} show the accuracy, macro $F_{1}$ scores, multiclass AUROC, and macro average AUPRC values for both tasks and for all of the classifiers, trained with and without ASR data and evaluated on human and ASR transcripts separately. They also show whether calibrated or uncalibrated models performed the best across all metrics. Tables \ref{soap-section-table} and \ref{speaker-label-table} break down the contributions of the different layers in the deep learning architecture for each SOAP section and speaker label.  For all metrics across both tasks, the deep learning models which had at least ELMo, layer attention, word level attention, and a bi-LSTM at the utterance embedding level performed the best.  These models with contextualized utterances improved SOAP section classification accuracy by 1\% compared to the best performing deep learning models without contextualized utterances and 1-2\% compared to the best performing BoW based models. Similarly, they improved macro $F_{1}$ scores by 6-7\% compared to simpler deep learning models, and 12\% compared to BoW models. AUROC increased by 9-10\% compared to simpler deep learning models by 13-15\% compared to BoW models. Finally, AUPRC increased by 7-9\% compared to simpler deep learning models and 9-13\% compared to BoW models. A similar pattern of gains was observed for the speaker label classification task. When breaking down the multiclass problems by SOAP section and speaker label, again models with contextualized utterances often resulted in improvements to the $F_{1}$ score of 10\% or more, with the exception of the majority class--in which case contextualizing utterances had very little effect on classification $F_{1}$ scores. On average across all SOAP sections, our models performed better than previously reported results for SOAP classification \citep{jeblee2019extracting}, with particularly large improvements to the lower frequency classes. We achieve 5-18\% improvements in $F_{1}$ scores for the Objective, Assessment, and Plan classes, while our model achieves a 2\% lower $F_{1}$ score on the Subjective.  Finally, the macro $F_{1}$ score achieved by our best performing model was similar to the $F_{1}$ score achieved by using two independent human annotations as ground truth and test examples (see Table \ref{qc-analysis-table2} in Appendix \ref{quality-control-appendix}.

As for the effect of calibration, calibrated models generally performed as well as or better than uncalibrated models on measures of accuracy and AUPRC. For AUROC, uncalibrated models typically outperformed calibrated models. $F_{1}$ scores, on the other hand, show a more complex interaction with model calibration. In Tables \ref{soap-section-table} and \ref{speaker-label-table}, it appears that uncalibrated models outperformed calibrated models for classes which were relatively infrequent, but calibrated models outperformed uncalibrated models for classes which were more frequent. Given that calibrated models performed as well as or better than uncalibrated models on AUPRC, it may also be the case that, with optimal thresholds, calibrated models could achieve similar $F_{1}$ scores to uncalibrated models. It should also be kept in mind that the models' loss functions were not designed to optimize $F_{1}$. Still, the possibility that calibration may have adverse effects on a model's ability to detect rare events should be investigated more thoroughly in future research.

Finally, we found mixed evidence for the use of ASR in training. When evaluating on the ASR test set, the best performing models trained on ASR data as opposed to the best performing models trained only on human transcripts often had similar or improved accuracy, $F_{1}$, AUROC, and AUPRC scores. By task, the effect of including ASR in training was more noticeable for speaker classification than for SOAP section classification. For SOAP classification, there was little to no systematic effect, while including ASR in training improved all measures for speaker classification. Overall, these improvements were smaller than the effect of contextualizing utterances.

\section{Discussion} 
\label{discussion}

\subsection{Natural Language Understanding for Medical Conversations}
\label{discussion-natural-language-understanding-of-clinical-dialoguues}

The deep learning architecture we proposed achieved the best performance on speaker label and SOAP section classification of utterances from medical conversations. Because of the novelty of the SOAP section classification task, it is somewhat difficult to know what constitutes objectively good or poor performance on SOAP section classification. On the one hand, this can be seen as a strength as the current analysis fills an important gap in research on medical conversations. On the other hand, it precludes our ability to assert that the performance reported in the results is state-of-the-art. 

Fortunately, we do have a reference for human performance on this task as part of the Quality Control analysis presented in Appendix \ref{quality-control-appendix}. This should be taken with a grain of salt since the quality control dataset was different from the test set used to evaluate model performance. Nevertheless, when we use one set of human annotations for a given medical conversation to predict which utterances would be assigned to which SOAP sections by another independent annotator we achieve a macro $F_{1}$ of 0.47, which is 1\% lower than our best performing model. Again, this can be seen as both good and bad. Our best models achieve near human performance. However, the way we have formulated the problem in the current paper means that we there is little room for further improvement.

The relatively low agreement between annotators is difficult to interpret because it does not necessarily mean that the annotators wrote substantially different SOAP notes, only that they cited different evidence. Citing evidence is an inherently subjective task; two humans are unlikely to agree 100\% on what constitutes the best set of evidence for a specific claim. Aside from that, clinical conversations can be highly repetitive. For example, an individual with diabetes may make numerous references to their diabetes throughout a medical conversation with their healthcare provider. Thus, there may be numerous pieces of non-overlapping but essentially equivalent pieces of evidence for a single clinical note. This means that downstream tasks may benefit from multi-evidence question answering architectures \citep{Zhao2020Transformer-XH:}.

Perhaps even more interesting were the results showing which layers in our deep learning architecture contributed the largest gains to overall performance. All of the deep learning models outperformed the BoW models on all metrics. This, unsurprisingly, shows that models which capture linguistic structure generally outperform those that do not. Furthermore, the inclusion of a bi-LSTM at the utterance level led to substantial improvements across multiple measures regardless of task or training and test conditions. This suggests that for medical conversations, understanding who is speaking and what is or is not clinically significant are highly contextual tasks. This has important implications for the use of machine learning for understanding medical conversations because many approaches currently rely on ontology based tools such as those provided as part of the Unified Medical Language System (UMLS), string matching techniques, or may use context independent word embeddings \citep{finley2018automated,aronson2010overview,bodenreider2004unified,mikolov2013efficient, mikolov2013distributed}. These models, like our deep learning models which included word level attention, may be good at capturing important vocabulary or even intrasentential context if parsing is used, but they would not be able to capture intersentential context like our models which use a bi-LSTM at the utterance evel. Though intrasentential context techniques may work well for highly structured EHR data or well formulated prose, our results suggest that they are suboptimal for medical conversations between doctors and patients.

\subsection{Mapping Human Clinical Annotations to ASR Output}

In this paper we explore a modular method for noisy label mapping of clinical annotations to ASR data. The method relies on a combination of string alignment algorithms and label smoothing to account for the additional noise. Using this mapping to include data in training improved classifier performance on ASR transcripts only for the speaker classification task, but not the SOAP section classification task. Thus, the effects of including ASR data in training may be more task dependent and overall more modest than the effects of including contextualized utterances. In particular, the null finding for SOAP classification may be due to the fact that the model trained only on human transcribed data was already at ceiling--or near human performance--meaning that there was no room for improvement by adapting the model to ASR output. Thus, human tasks which have relatively moderate or low inter-rater reliability may not benefit as much from adaptations to ASR noise due to the large amount of noise already present in the data. 

The primary factor which differentiates our method from others is modularity. Most other methods use multimodal representations and joint, end-to-end training to achieve robustness to ASR output \citep{Serdyuk2018TowardsES, lakomkin2019incorporating, morbini2012reranking}. In practice, this means that these models must train ASR and NLU architectures at the same time to realize the benefits. This increases the complexity of training in terms of time, total number of parameters in the model, and computer memory. It can also make it more difficult to troubleshoot poor model performance since there are more variables and interactions that could be causing problems. Another appealing aspect of our modular method is that it can be applied to publicly available state-of-the-art ASR systems like the Google model used in this paper. Of course, we fully acknowledge the contributions and potential of these less modular approaches, and the choice of which method is best will depend on the goals of and resources available to a specific research group. Our argument is more about the benefits of modularity in general than any specific aspects of the previous research.  

\subsection{Limitations}
\label{limitations}

Ultimately, we are concerned with whether the evidence that our model finds is ``complete" or not, in the sense that it is sufficient to generate an acceptable quality SOAP note. We are not explicitly concerned with whether it cites exactly the same evidence as one human would. Thus, the problem formulation in the current study does not provide directly interpretable quantitative data on how an end-to-end SOAP note solution would perform in practice. Still we believe that the results are promising and provide a foundation for future research. Future work should focus on information extraction and summarization of the various SOAP sections and subsections, which would have more interpretable results for practical applications of end-to-end systems. 

Also, there are several aspects of the current research which could not be rigorously analyzed in the scope of this paper. These include the mapping technique which we presented and the multitask architecture. As for the former, we know that there are areas where the mapping needs to be improved. For example, the diarization, and consequently sentence segmentation, was particularly prone to problems during periods of dense turn taking with short sentences (e.g. the patient and doctor going back and forth saying things like ``Ok.", ``Good.", ``Mhm.") and when one speaker would briefly interject extended speech by the other speaker (e.g. when the patient periodically says ``Ok." to indicate that they are listening when the doctor talks for an extended period of time). A detailed discussion of this mapping can be found in section \ref{mapping-human-annottaions-to-ASR-output}. We believe that this is one primary source of inconsistencies between the utterance segmentation of the ASR data and the human segmented utterances in the human transcripts. Segmentation related problems would likely benefit from multimodal models which can use both acoustic and prosodic evidence from audio data as well as linguistic evidence from text. As for the multitask architecture and the choice of task in general, there are many possible relevant tasks which could be used to measure basic understanding of medical conversations. Future research should explore a larger array of tasks and examine whether solving them independently or simultaneously in a multitask framework has any benefits for model performance.

\begin{acks}
We thank: University of Pittsburgh Medical Center (UPMC) and Abridge AI Inc. for providing access to the de-identified data corpus; our colleagues: Sai P. Selvaraj, Nimshi Venkat Meripo, Anirudh Mani and Dr. Shivdev Rao for insightful discussions; Steven Coleman, Deborah Osakue, and Stephanie Shields for data business development and annotation management.
\end{acks}

\small
\bibliography{main}

\appendix
\section{Quality Control}
\label{quality-control-appendix}

\subsection{Quality Control Dataset}

We also collected a dataset of 599 medical conversations which were SOAP annotated by two independent annotators. We used these pairs of independent SOAP notes to evaluate inter-rater reliability of the data used in this study. The dataset contained notes from over 100 different annotators. It should be noted that this dataset was different from the test set used for evaluation in the results (section \ref{results}) of the main text. Though it is a subsample of the same larger dataset, it differs primarily in that it is not a true random sample of the data. Due to resource constraints, the files used in this quality control dataset were all between 4.5 and 5.5 minutes in length, which only spans a small portion of the actual variation in encounter length. The test set used in the main text, however, does sample from all possible encounter lengths.

\subsection{Evaluation Metrics}

Because there is no standardized way to measure inter-rater reliability for a SOAP note, we created our own metrics as well as target numbers which the annotators were expected to meet. The concept for the metrics was borrowed from the analysis of ASR systems (and DNA sequencing), in which every note was categorized as identical, a deletion, an insertion, or a substitution. Given two SOAP notes, a reference SOAP note and a source SOAP note, every note from the source SOAP note was mapped to the reference SOAP note as follows: a given note in the source was mapped to that note in the reference which was placed in the same SOAP subsection and which had the highest overlap in cited evidence and tags. If a note in the source note could not be mapped to any note in the reference (due to lack of overlap in tags or evidence), it was categorized as an insertion. Conversely, if a note in the reference could not be mapped to any note in the source, it was considered a deletion. Notes which were successfully mapped but were not identical were categorized as substitutions, and notes which were perfect matches were categorized as identical. Because an identical note had to have the exact same tags, short summary, and cited evidence, it was exceedingly rare for notes to be categorized as identical. 

We set targets for the annotators to have no more than 30\% deletions and no more than 30\% insertions for all notes in the SOAP note except for two sections, a miscellaneous subsection and the assessment. These two sections were highly unstructured and, consequently, had the lowest inter-rater reliability. In addition, we set criteria for the degree of overlap for notes which were categorized as substitutions, to ensure that notes which were being mapped were actually highly similar. These targets were 70\% overlap in cited evidence, and 70\% overlap in tags for all substitutions in a SOAP note. 

In addition to the alignment metrics, we report metrics which are more easily compared to traditional machine learning and clinical analyses. These metrics include, accuracy and f1 (the harmonic mean of precision and recall, also known as sensitivity) over all classes (macro f1) and for individual classes. For these metrics, we use one set of human annotations as the ``ground truth" and one as the ``predicted label." These metrics provide a human baseline for measuring classifier performance, although, again, we note that the test set is based on different SOAP notes than the test set used for evaluating classifier performance. We also present the prevalence (prior probability) of different SOAP note classes across both sets of annotations, the conditional probability that one human annotator will mark a particular utterance as relevant to one of the four SOAP note sections given that the other human annotator also did, and the conditional probability that one human annotator will mark a particular utterance as relevant to one of the four SOAP note sections given that the other human annotator did not. 

\subsection{Results}

Table \ref{qc-analysis-table} displays the results for note level alignment statistics: deletions, insertions, identical, substitutions, tag overlap, and evidence overlap. The averages and variance from 599 files are shown. The annotators met the target metrics that we set for them in all cases, meaning that the rates of deletions and insertions were not significantly higher than 30\% on average, and evidence and tag overlap for substitutions were not significantly less than 70\%. Table \ref{qc-analysis-table2} provides a human baseline for accuracy and f1, and also reports the prevalence and odds of multiple annotators classifying utterances into different SOAP sections. When assess all the sections combined, accuracy was around 54\%, while macro ff1 was around 0.47. Individual section accuracies were mostly driven by the class prevalence, indicating the accuracy may not be a good metric for understanding classifier performance on the individual SOAP sections. F1 scores were 0.51 for the Subjective and Objective, 0.38 for the Plan, and 0.30 for the Assessment. This suggests that the annotators were mostly likely to disagree about cited evidence for the Plan and Assessment. Still, a human annotator citing an utterance as evidence to support a note written in one of the for SOAP sections was a reasonably strong indicator that another annotator would also do so. For sections with low prevalence like the Plan and Objective, an annotator citing an utterance as evidence for a particular SOAP meant it was anywhere from around 10-30 times more likely that the other annotator also would cite it as evidence for the same SOAP section, compared to when one annotator did not use a particular utterance as evidence.

\begin{table}
  \small
  \caption{Using Alignment Statistics to Evaluate Inter-Rater Reliability}
  \label{qc-analysis-table}
  \centering
  \begin{tabular}{lcccccc}
    \toprule
    \cmidrule(r){1-2}
    & & & & & Evidence & Tag \\
    & Identical & Del & Ins & Sub & Overlap & Overlap \\
    \midrule
    Mean & 0.008 & 0.304 & 0.256 & 0.687 & 0.694 & 0.83 \\
    Variance & 0.001 & 0.034 & 0.026 & 0.034 & 0.014 & 0.013 \\
    \bottomrule
\multicolumn{7}{p{.8\textwidth}}{Notes. Del = Deletions; Ins = Insertions, Sub = Substitutions. Evidence and tag overlap were only measured for substitutions.} 
\end{tabular}
\end{table}

\begin{table}
  \small
  \caption{Using Accuracy, F1, Prevalence, and Odds to Evaluate Inter-Rater Reliability}
  \label{qc-analysis-table2}
  \centering
  \begin{tabular}{lccccc}
    \toprule
    \cmidrule(r){1-2}
    Domain & Accuracy & F1 & Prevalence & $P(Y_{1}=1|Y_{2}=1)$ & $P(Y_{1}=1|Y_{2}=0)$ \\
    \midrule
    All Sections & 0.54 & 0.47* & -- & -- & -- \\
    None & 0.63 & 0.65 & 0.54 & 0.66 & 0.41 \\
    Subjective & 0.77 & 0.51 & 0.24 & 0.51 & 0.15 \\
    Objective & 0.97 & 0.51 & 0.03 & 0.51 & 0.02 \\
    Assessment & 0.72 & 0.30 & 0.20 & 0.31 & 0.17 \\
    Plan & 0.92 & 0.38 & 0.06 & 0.38 & 0.04 \\
    \bottomrule
\multicolumn{6}{p{.95\textwidth}}{Notes. *F1 for all sections is the macro f1 score. $Y_{1}$ and $Y_{2}$ refer to the two independent annotations of the same medical conversation.} 
\end{tabular}
\end{table}

\end{document}